\PassOptionsToPackage{numbers, compress}{natbib}

\documentclass{article}
\usepackage{booktabs}
\usepackage{adjustbox}

\usepackage[final]{neurips_2023}

\usepackage{listings}
\usepackage{xcolor} 

\usepackage[hyphens]{url}
\usepackage{hyperref}
\hypersetup{breaklinks=true}

\usepackage{natbib}
\usepackage{tabularx}
\usepackage{multirow}

\usepackage[utf8]{inputenc} 
\usepackage[T1]{fontenc}    
\usepackage{hyperref}       
\usepackage{url}            
\usepackage{booktabs}       
\usepackage{amsfonts}       
\usepackage{nicefrac}       
\usepackage{microtype}      
\usepackage{xcolor}         

\usepackage{graphicx}   
\usepackage{caption}    
\usepackage{subcaption} 

\title{The Hidden Structure - Improving Legal Document Understanding Through Explicit Text Formatting}

\author{%
  Christian Braun\thanks{Equal contribution. Listing order is alphabetical.}, \quad
  Alexander Lilienbeck\footnotemark[1], \quad
  Daniel Mentjukov\\
  Osborne Clarke\\
  \texttt{\{christian.braun, alexander.lilienbeck, daniel.mentjukov\}@osborneclarke.com}
}

\begin{document}

\maketitle

\begin{abstract}
The integration of \textbf{Large Language Models (LLMs)} into legal practice presents both transformative opportunities and significant risks, underscoring the need for rigorous benchmarking and a nuanced understanding of factors influencing their performance. Legal documents, particularly contracts, possess an inherent, semantically vital structure (e.g. sections, clauses) that is crucial for human comprehension but whose impact on LLM processing remains under-explored. This paper investigates the effects of explicit input text structure and prompt engineering on the performance of two LLM versions (GPT-4o and GPT-4.1, via Microsoft Azure) on a legal question-answering task using an excerpt of the \textbf{Contract Understanding Atticus Dataset (CUAD)}. We compare model exact-match accuracy (EM-Acc.) across various input formats: original well-structured plain-text (human-generated from CUAD), plain-text cleaned of line breaks, extracted plain-text from Azure OCR, plain-text extracted by GPT-4o Vision, and extracted (and interpreted) Markdown (MD) from GPT-4o Vision. To give an indication of the impact of possible prompt engineering, we assess the impact of shifting task instructions to the system prompt and explicitly informing the model about the structured nature of the input (e.g., "You will receive structured Markdown // Plain Text"). Our findings reveal that GPT-4o exhibits considerable insensitivity to variations in input structure but shows limited overall performance. Conversely, GPT-4.1's performance is highly sensitive to structure; poorly structured inputs yield suboptimal results (but identical with GPT-4o), while well-structured formats (original CUAD text, GPT-4o Vision text and GPT-4o MD) improve exact-match accuracy by approximately 20 percentage points. Optimizing the system prompt to include task details and an advisory about structured input further elevates GPT-4.1's accuracy by an additional ~10-13 percentage points, with Markdown ultimately achieving the highest performance under these conditions (79 percentage points overall exact-match accuracy). This research empirically demonstrates that while commonly used models exhibit greater resilience, careful input structuring and strategic prompt design remain critical for optimizing the performance of models like GPT-4.1, significantly influencing outcomes in high-stakes legal applications. We highlight the often-overlooked impact of such "minor" variations, urging critical scrutiny of all factors shaping LLM behavior in sensitive domains.
\end{abstract}


\section{Introduction}

The rapid expansion of Large Language Models (LLMs) offers transformative potential for numerous sectors, not least the field of law, where their integration has the potential to enhance—and in some cases fundamentally alter—established methods of legal practice and analysis. Yet, this growing enthusiasm is necessarily tempered by a clear recognition of the substantial risks associated with deploying such powerful, yet incompletely understood, technologies into the legal domain \cite{gelbach2021}. Existing research has thoroughly documented the tendency of LLMs to produce content that may be offensive, misleading, or factually inaccurate \cite{bender2021, liang2023}. The potential for such errant behaviors to arise within legal applications \cite{romoser2023} raises legitimate concerns regarding substantial societal harms \cite{weiser2025}, with a disproportionate burden potentially falling upon historically marginalized and under-resourced communities likely to bear a disproportionate share of the impact \cite{surden2020, volokh2023}. Accordingly, the need for robust infrastructure and rigorous methodologies to benchmark LLM performance and ensure their safe deployment in legal contexts is both urgent and essential.

Despite the urgency of this need, both practitioners and researchers face significant obstacles in evaluating the ability of LLMs to execute complex legal reasoning tasks or furnish verifiably accurate responses to legal queries \cite{legalbench2023}. A critical yet often underexamined aspect of this challenge lies in the structural characteristics of legal texts. Legal documents—contracts in particular—are not unstructured bodies of prose but are systematically organized through a formal hierarchy of titles, sections, subsections, clauses, and enumerated provisions \cite{wang2023}. This formal organization is not merely a matter of stylistic convention; rather, it carries profound semantic import, defining the scope of agreements, articulates obligations, outlines exceptions, and ultimately governs legal interpretation and enforcement.

While contemporary LLMs have demonstrated impressive capabilities in processing vast quantities of natural language, the degree to which their analytical performance depends on— or is enhanced by—explicit structural cues in the input text remains insufficiently understood, particularly in the complex and high-stakes context of legal analysis \cite{lai2023}. While it is often assumed that these models can implicitly infer underlying structure from unformatted text \cite{tan2024}, however, the validity of this assumption may not extend uniformly across different model architectures or generations. Moreover, this capability cannot be reliably presumed when input texts are degraded, whether due to errors introduced by Optical Character Recognition (OCR) or through overly aggressive pre-processing aimed at standardizing heterogeneous sources \cite{hamad2016, hegghammer2022, nguyen2022}.

In this paper, we endeavor to contribute to this critical area of inquiry through a focused empirical investigation. Our approach is primarily twofold:

\begin{enumerate}
    \item We examine the hypothesis (\textbf{H1}) that the explicit structure of input text—ranging from comparatively unstructured OCR output to well-formatted plain text (TXT) and syntactically rich Markdown (MD)—materially affects the performance of two commonly utilized LLM versions deployed on the Microsoft Azure platform (GPT-4o and GPT-4.1) when applied to a standardized legal question-answering task (excerpt of CUAD dataset).
    \item We investigate the impact of specific, but simple prompt engineering strategies (\textbf{H2}), particularly the provision of meta-information to the model regarding the structured nature of its input (e.g., explicitly conveying "You will receive structured Markdown // Plain Text").
\end{enumerate}

We evaluate these hypotheses empirically using an excerpt of the tasks of the Contract Understanding Atticus Dataset (CUAD) \cite{hendrycks2021}, a widely used benchmark, assessing performance based on the model’s accuracy in identifying and citing relevant contractual clauses—a core competency in legal practice.

The principal contributions of this study, therefore, can be summarized as follows:

\begin{itemize}
\item We present a systematic, comparative analysis of GPT-4o and GPT-4.1 performance across a spectrum of input text structures derived from authentic legal documents, thereby offering insights into model-specific sensitivities to data representation.
\item We furnish empirical evidence demonstrating significant performance enhancements for the GPT-4.1 model when supplied with appropriately structured input / text formatting.
\item We offer a nuanced elaboration of the intricate interplay between input format (specifically, plain TXT versus Markdown), model version, and potential prompt design (or marginal changes to prompt design) within the specialized context of legal document analysis, contributing to a more granular understanding of factors influencing LLM efficacy in this domain.
\end{itemize}

By illuminating these dynamics, this research seeks to inform best practices for preparing legal texts for LLM processing and to emphasize the enduring importance of careful data curation / structuring texts and also minor changes in prompt design, even as language models continue to advance in capability.

Before progressing further, we emphasize that the purpose of this work is to draw attention to the often-overlooked influence of seemingly minor variations between model versions—particularly in contexts where prevailing assumptions may obscure meaningful differences, such as the impact of structured text formatting in newly released LLMs. Legal practitioners and organizations should be aware that small modifications in input preparation, prompt design, or even the technical implementation of prompts across different provider interfaces (e.g., OpenAI, Google, Microsoft) can yield substantial effects on model performance. With this study, we aim to initiate a broader discourse around the need for critical scrutiny of the many factors—however subtle—that shape LLM behavior in high-stakes applications like law.

\section{Background and Related Work}
\label{gen_inst}

The effort to employ Large Language Models (LLMs) for complex analytical tasks—particularly in domains requiring high precision, such as legal document analysis—demands close attention to the manner in which information is conveyed to these models \cite{wang2024, shen2025}. The interplay among input structure, prompt formulation, and model performance represents a crucial focal point for both current research and practical refinement.

Legal documents are inherently structured. This study adopts Markdown as the primary format for representing such structure to large language models (LLMs), based on several converging considerations. First, Markdown, by its design, provides a lightweight yet expressive syntax for articulating document hierarchy (e.g., headings), emphasis, lists, and other structural elements that are often crucial for semantic disambiguation in legal texts \cite{euguidelines2025}. This capacity to transparently transport structural information is paramount \cite{hegel2021}. Second, this choice aligns with emerging best practices recommended by leading LLM developers, including OpenAI, who advocate for clearly structured inputs, with Markdown frequently cited as an effective format \cite{openai2025}. Third, prior research, though often domain-general, has provided preliminary indications that structured input formats, including Markdown, can contribute to enhanced performance in various LLM-driven tasks by making salient features more explicit \cite{tan2024}.

The basis for evaluating legal documents is built on large, expert-annotated resources.
\textsc{LegalBench} aggregates 162 tasks spanning statutory interpretation, case outcome prediction, and contract review, establishing a community standard for legal-reasoning measurement \cite{legalbench2023}.
For contract analysis specifically, \textsc{CUAD} provides 13 k span-level annotations across 41 clause types and continues to serve as a stress-test for LLM comprehension of real contracts \cite{hendrycks2021}. On this dataset, “\textit{Better Call GPT}’’ reports that GPT-4 reaches or exceeds senior-lawyer accuracy while reducing review time from hours to seconds and cost by two orders of magnitude \cite{martin2024}. Our study complements these works by dissecting how input cleanliness and prompt placement mediate such gains. The \textsc{MDEval} benchmark quantifies “Markdown-awareness’’ across 20 k instances and finds that models scoring higher on its hierarchy-sensitive tasks also yield better factual accuracy in downstream QA and reasoning settings \cite{chen2025}.
Analogous effects have been observed in other safety-critical domains such as medicine, where a two-step “structured clinical-reasoning’’ prompt raised diagnostic top-3 accuracy from 66.5\% to 70.5\% \cite{sonoda2024}. These findings align with our choice of Markdown: it preserves headings, clause boundaries, and list semantics while remaining token-efficient—properties that neural models appear to exploit.

The articulation of tasks to the LLM—commonly referred to as “prompt engineering”—has become a key factor influencing output quality and relevance \cite{sahoo2025}. The initial phase of our experimental design involved a baseline setup wherein task instructions were embedded within the user prompt, leaving the system prompt largely generic (e.g., "You are a helpful assistant"). This approach, common in many applications, served as a control condition for evaluating the effects of more sophisticated prompting strategies.
\textsc{The Prompt Report} surveys 58 prompting techniques and distils best-practice guidelines for state-of-the-art models \cite{schulhoff2025}.  
He et al.\ systematically vary whitespace, bulleting, and role headers, showing up to 40\% accuracy swings on code-translation tasks for GPT-3.5 while GPT-4 is more robust but still affected \cite{he2024}. \textsc{PromptBench} and the NeurIPS 24 “PromptEval’’ protocol further reveal that evaluation results can flip when only the prompt template, not the model, is changed \cite{zhu2024,polo2024}. 
Our experimental manipulation of system- versus user-prompt placement extends this line of inquiry to the legal QA setting.

The importance of prompt architecture has been increasingly emphasized in recent research. For instance, He et al. (2024) in "Does Prompt Formatting Have Any Impact on LLM Performance?" \cite{he2024} explore the empirical impact of various prompt structures, finding that formatting choices can indeed yield differential performance outcomes, often in model-specific ways. This resonates with observations from other researchers which examine the nuanced effects of prompt formatting across different LLM architectures, see e.g. \cite{han2025}. Additionally, guidance from organizations such as OpenAI — including resources like the OpenAI Cookbook and model-specific prompting guides for GPT-4.1 \cite{openai2025} — consistently highlights that structured prompts, with clearly defined roles, contextual framing, and explicit output formats, can substantially improve an LLM’s ability to manage complex instructions and large-scale textual inputs \cite{opengpt42024}.

Preliminary investigations conducted prior to the formal experiments reported in this study revealed patterns that directly shaped our hypotheses and experimental design.

\section{Experiments}
\label{headings}

To empirically investigate our hypotheses regarding the impact of input text structure and prompt engineering on the performance of Large Language Models in the legal domain, we designed a series of controlled experiments. This section details the experimental setup, including the dataset, task definition, models employed, input modalities, prompting strategies, and evaluation metrics. Subsequently, we will present the experimental procedure and the results obtained.

\subsection{Experimental Setup}

Our experimental design aims to systematically isolate and quantify the effects of (i) varying levels of explicit structural information in the input legal text and (ii) different prompt engineering paradigms on the question-answering accuracy of selected LLMs.

\subsubsection{Dataset and Task Definition}
The empirical investigation leverages the \textbf{Contract Understanding Atticus Dataset (CUAD)} \cite{hendrycks2021}. For our experiments, we focused on a core legal question-answering task inherent to contract review. We used a proportion of CUAD to validate our assumptions (see \ref{appendix:A}).\footnote{Our decision has been taken to extract a sample from the comprehensive data set, with the objective of providing an indication without incurring undue runtime, financial and effort costs.} After filtering the relevant topics (in this case: "Competitive Restriction Exception", "Non-Compete", "Exclusivity", "No-Solicit of Customers") 928 questions were used for each run (after running through the predefined template - see \ref{appendix:A}). Additionally, we extracted the same questions with their paired answers from LegalBench-RAG \cite{pipitone2024} to complement all needed information.  

The task requires the LLM to process a given contract and answer a specific question pertaining to its clauses. The expected output is bipartite:
\begin{enumerate}
\item A categorical response ("Yes" or "No") indicating the presence or absence of the condition queried.
\item If the condition is present, an exact verbatim citation of the contractual clause that substantiates the "Yes" answer (otherwise it will be stated as "No").
\end{enumerate}

This two-part task structure rigorously tests not only the model's ability to identify relevant information but also its precision in extracting specific textual evidence, a critical function in legal due diligence and analysis (see \ref{appendix:B}).

\subsubsection{Language Models}
We evaluated two distinct OpenAI's Generative Pre-trained Transformer (GPT) model series, accessed via the Microsoft Azure OpenAI Service:
\begin{itemize}
    \item \textbf{GPT-4o:} This denotes the "omni" model from OpenAI, representing their then-latest generation multimodal model, purported to offer enhanced capabilities, efficiency, and potentially different sensitivities to input variations \cite{openaigpt4o}.
    \item \textbf{GPT-4.1:} This refers to a version of the GPT-4 series available on Azure prior to the widespread release of GPT-4o or similar iterative updates to the core GPT-4 architecture. It represents a highly capable, but potentially more established, iteration \cite{openaigpt4.1}.
\end{itemize}

The selection of these two model series allows for a comparative analysis across different points in the LLM development trajectory, potentially revealing evolving model characteristics concerning input structure robustness and prompt adherence.

\subsubsection{Input Text Preparation and Modalities}
\begin{figure}[htbp]
  \centering
  \begin{subfigure}[t]{0.32\textwidth}
    \centering
    \includegraphics[width=\linewidth]{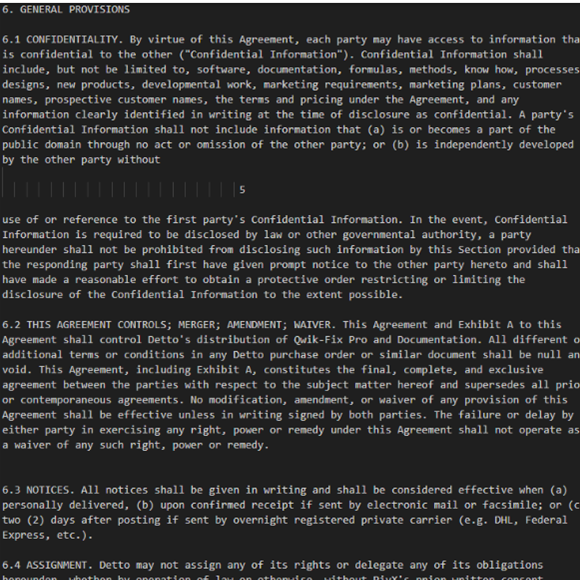}
    \caption{\texttt{.txt} (Original CUAD-Dataset)}
    \label{fig:cuad}
  \end{subfigure}
  \hfill
  \begin{subfigure}[t]{0.32\textwidth}
    \centering
    \includegraphics[width=\linewidth]{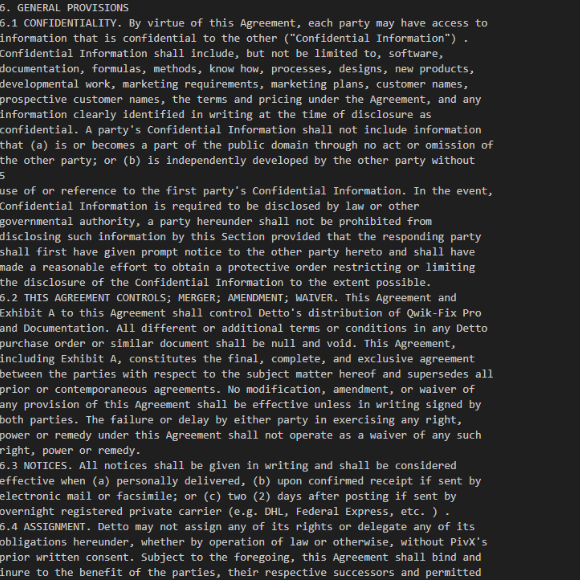}
    \caption{\texttt{.txt} (Azure OCR)}
    \label{fig:azure}
  \end{subfigure}
  \hfill
  \begin{subfigure}[t]{0.32\textwidth}
    \centering
    \includegraphics[width=\linewidth]{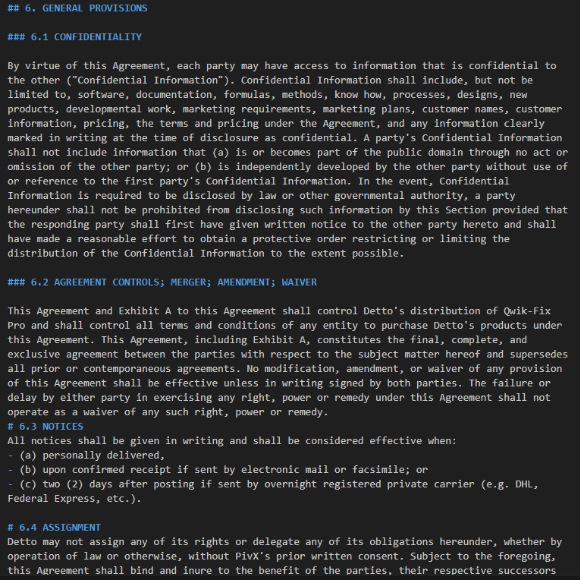}
    \caption{\texttt{.md} (GPT-4o PDF-vision-parse)}
    \label{fig:gpt4o}
  \end{subfigure}

  \caption{Visual comparison of file formats
 of the same contract section.}
  \label{fig:vergleich}
\end{figure}

To investigate the impact of input text structure (i), the contract content was prepared and presented to the LLMs in five distinct modalities, representing a spectrum of structural explicitness and data fidelity:

\begin{enumerate}
    \item \textbf{Original TXT from CUAD Dataset (Well-Structured - $'CUAD_OrigTXT'$)}: These are the original plain text files directly from the CUAD dataset. These texts generally exhibit good inherent structure, characterized by appropriate paragraph breaks, indentation, and spacing that visually delineate sections and clauses, reflecting human-readable formatting. In Figure 1 \ref{fig:cuad} is displaying an example of the Original CUAD-Dataset.
    \item \textbf{Original TXT from CUAD Dataset (Cleaned with Regex - $'CUAD_RegexCleanTXT'$)}: The $`CUAD_OrigTXT`$ files were subjected to a regular expression-based cleaning process designed to remove what might be considered "excessive" or multiple consecutive line breaks. The intention was to create more compact text blocks, potentially obscuring some of the visual and implicit structural cues present in the original files. This simulates an aggressive, potentially information-degrading, normalization step resulting in an unstructured format (and therefore more real-world case setting). The primary goal was to exclude OCR or recognition errors as a confounding factor, thereby isolating the difference between structured and unstructured input as the sole source of variation in the observed outcomes.
    \item \textbf{TXT from Image PDF with Azure OCR ($`AzureOCR_TXT`$)}: This modality simulates a common real-world scenario where contracts are available only as scanned image PDFs. The text was extracted using Microsoft Azure Cognitive Services for Vision (OCR) \cite{microsoftOCR}. This process is known to be imperfect, potentially introducing character recognition errors, layout misinterpretations, and a general loss of fine-grained formatting. Figure 1 \ref{fig:azure} is showing an example after using an image-created version of the original pdfs of the CUAD-Dataset and running through Microsofts Azure OCR Engine.
    \item \textbf{TXT from Image PDF with GPT-4o Vision ($`GPT4oVision_TXT`$)}: Contract text was hypothetically extracted from image PDFs utilizing the vision capabilities of GPT-4o itself, configured to output plain text. This modality aims to assess the quality of text extraction from a state-of-the-art vision-language model, potentially offering better structural preservation than traditional OCR but still in a plain text format. We used the GPT-4o model with vision capabilities (using base64 encoding to "prompt" the related image).
    \item \textbf{MD from Image PDF with GPT-4o Vision ($`GPT4oVision_MD`$)}: Similar to the above, but GPT-4o Vision was prompted to generate Markdown (MD) output from the (hypothetical) image PDFs. For this, we used the github library vision-parse for applying different markdown strategies (see \cite{visionparse}). The objective here was to leverage GPT-4o's multimodal understanding to not only extract text but also explicitly encode structural elements (e.g., headings, lists, emphasis) using Markdown syntax, thereby providing the most explicit structural cues to the downstream LLM. This is shown in Figure 1 \ref{fig:gpt4o}, where the Original PDFs from the CUAD-Dataset are running through vision-parse \cite{visionparse} using GPT-4o with vision capabilities and prompting it to generate clean, structured MD-format.
\end{enumerate}

Setting vision-parse we enabled parallel processing of multiple pages (enable\_concurrency=True) and detailed extraction Enable for retrieviong complex information such as LaTeX equations, tables, images (detailed\_extraction=True).\footnote{All standard settings and best-practices can be found in the vision-parse github lib. \cite{visionparse}} For downstream language model processing, we set the temperature parameter to 0.2—favoring more deterministic outputs given the legal domain context—and imposed no explicit limit on the maximum number of tokens.

This diverse set of input modalities allows for a systematic evaluation of model performance as a function of the richness and accuracy of structural information conveyed in the input contract.

\subsubsection{Prompting Strategies}
We employed two primary prompting strategies (ii) to assess the influence of how task instructions and context are provided to the LLMs:

\paragraph{Strategy 1:} User-Centric Task Definition ($`S1_UserTask`$)

This strategy places the primary task instructions within the `User` message.

\begin{lstlisting}
SYSTEM_PROMPT = """
You are a Helpful assistant.
"""

USER_PROMPT = """
Answer the following question in two parts: (1) Yes/No for the first condition, 
(2) cite the exact wording of the found clause.

# Question
Here is the question:
{question}

# Contract
Here is the contract:
{contract_content}
"""
\end{lstlisting}

This configuration represents a common approach where the system prompt is generic, and the user prompt carries the full task specification alongside the data.
\newpage
\paragraph{Strategy 2:} System-Centric Task Definition with Structure Awareness ($`S2_SystemTaskStruct`$)

This strategy relocates the core task instructions to the `System` message and explicitly informs the model about the nature of the incoming structured data.

\begin{lstlisting}
SYSTEM_PROMPT = """
You will receive a contract in a Markdown // Plain Text format.

Answer the following question in two parts: (1) Yes/No for the first condition, 
(2) cite the exact wording of the found clause.

You receive structured Markdown // Plain Text 

*(Note: The segment "Markdown // Plain Text" was varied to accurately reflect the input type, e.g., "structured Plain Text" for TXT inputs and "structured Markdown" for MD inputs, to provide precise meta-information).*


USER_PROMPT = """
# Question
Here is the question:
{question}

# Contract
Here is the contract:
{contract_content}
"""
\end{lstlisting}

This strategy aims to leverage the system prompt for persistent instructions and prime the model to specifically anticipate and utilize the structural characteristics of the provided contract content.

These two strategies allow us to disentangle the effects of input formatting from the effects of how the model is instructed and primed for the task (and especially the impact using marginal changes in handling the API with SYSTEM\_PROMPT and constructing USER\_PROMPT).

\subsubsection{Evaluation Metrics}
The performance of the LLMs on the defined task was quantified using \textbf{exact-match accuracy} \cite{deepsetmetrics}. For an answer to be considered correct, both parts of the bipartite response must be accurate:
\begin{enumerate}
    \item The "Yes/No" component must correctly reflect the presence or absence of the queried condition in the contract.
    \item If the answer is "Yes," the cited clause must be an exact, verbatim match to the corresponding ground-truth clause in the CUAD annotations. No partial credit was awarded for near misses or semantically similar but non-identical citations. We gradually used an LLM-as-a-judge and \textit{answer\_correctness} to assess an evaluation on a content level \cite{llmasajudge, ragas2025}. \texttt{Note}: We didn't include the results in this study, as it should be published in a future report.
\end{enumerate}

\textit{Exact-match accuracy} is reported as the percentage of test instances for which the model produced a fully correct Yes/No answer. This stringent metric reflects the high precision required in legal applications.\footnote{Out goal was to evaluate the impact of the input - not building a benchmark for evaluating legal retrieval or content.}

\subsection{Results}

The experiments were conducted by systematically applying each prompting strategy across all defined input text modalities for both GPT-4.1 and GPT-4o. The primary metric, exact-match accuracy for the bipartite (Yes/No and clause citation) question-answering task, was recorded for each condition. The aggregated results are presented in Table 1 \ref{tab:results}.

\begin{table}[h!]
\centering
\caption{Exact-Match Accuracy (\%) of GPT-4.1 and GPT-4o on CUAD Question-Answering Task Across Varying Input Modalities and Prompting Strategies.}
\label{tab:results}
\begin{adjustbox}{max width=\textwidth}
\begin{tabular}{@{}llcc@{}}
\toprule
\textbf{Prompt Strategy} & \textbf{Input Text Modality} & \textbf{GPT-4o EM-Acc.} & \textbf{GPT-4.1 EM-Acc.} \\
\midrule
\multirow{5}{*}{\begin{tabular}[c]{@{}l@{}}\textbf{S1: User-Centric Task Definition}\\ (System: "You are a helpful assistant")\end{tabular}} 
& CUAD\_OrigTXT (Well-Structured) & 0.485736 & \textbf{0.677846} \\
& CUAD\_RegexCleanTXT (Cleaned) & 0.472384 &  0.481439\\
& AzureOCR\_TXT & 0.472748 & 0.489865\\
& GPT4oVision\_TXT & 0.474385 & \textbf{0.669834} \\
& GPT4oVision\_MD & 0.475639 & \textbf{0.668973} \\
\midrule
\multirow{3}{*}{\begin{tabular}[c]{@{}l@{}}\textbf{S2: System-Centric Task Definition}\\ \textbf{with Structure Awareness}\end{tabular}} 
& CUAD\_OrigTXT (Well-Structured) & 0.501491 & 0.775394 \\
& AzureOCR\_TXT & 0.511314 & 0.567569\\
& GPT4oVision\_MD & \textbf{0.523498} & \textbf{0.799760} \\
\bottomrule
\end{tabular}
\end{adjustbox}
\caption*{\footnotesize CUAD\_RegexCleanTXT and GPT4oVision\_TXT were excluded from Strategy S2 based on findings from Strategy S1, which demonstrated a significant performance advantage when utilizing structured data. The objective of Strategy S2 was to investigate the extent to which marginal modifications in prompting influence the structure and formatting of textual outputs. AzureOCR\_TXT was included in this analysis due to its widespread adoption in legal document processing, warranting an assessment of its responsiveness to these prompt-based variations.}
\end{table}

The results presented in Table 1 reveal several distinct patterns concerning the performance of GPT-4.1 and GPT-4o under varying conditions of input structure and prompt design. These findings are analyzed below in the context of our initial hypotheses.

\paragraph{Impact of Input Text Structure (Hypothesis H1)}

Our first hypothesis (H1) posited that providing more explicit structural information within the input text would improve LLM performance, particularly for models less inherently robust to input variations. The results from Prompting Strategy S1 (User-Centric Task Definition) provide substantial evidence supporting this hypothesis, especially for GPT-4.1.

\textbf{Performance of GPT-4.1 under Strategy S1}
\begin{itemize}
    \item When presented with input modalities characterized by poor or degraded structure ($`CUAD_RegexCleanTXT`$ and $`AzureOCR_TXT`$), GPT-4.1 achieved a relatively low accuracy of 48\%. This indicates a significant struggle to correctly perform the task when structural cues are minimal or corrupted.
    \item When GPT-4.1 received well-structured inputs—either the original, well-formatted CUAD texts ($`CUAD_OrigTXT`$) or texts extracted with better structural fidelity by GPT-4o Vision ($`GPT4oVision_TXT`$ and $`GPT4oVision_MD`$)—its accuracy markedly improved to 66-67\%. This represents a substantial gain of approximately 18-19 percentage points directly attributable to the enhanced structural quality of the input.
    \item Under this prompting strategy, there was no discernible performance difference for GPT-4.1 between high-quality plain text ($`CUAD_OrigTXT`$, $`GPT4oVision_TXT`$) and Markdown ($`GPT4oVision_MD`$), all achieving similar accuracy levels. This suggests that, without specific prompting to leverage Markdown's syntax, GPT-4.1 treated well-structured TXT and MD inputs comparably well.
\end{itemize}

\textbf{Performance of GPT-4o under Strategy S1:}
\begin{itemize}
    \item GPT-4o exhibited a strikingly different behavior. Its accuracy remained remarkably consistent across all five input modalities under Strategy S1, hovering between 47\% and 48\%. This demonstrates a high degree of robustness to variations in input text structure. The degradation of structure in $`CUAD_RegexCleanTXT`$ and $`AzureOCR_TXT`$ did not substantially impair its performance compared to the well-structured inputs.
    \item An important, though ancillary, observation is that GPT-4o's absolute accuracy scores under Strategy S1 were notably lower than those of GPT-4.1 when the latter was provided with well-structured input (47-48\% for GPT-4o vs. 66-67\% for GPT-4.1). This suggests that, for this specific task and evaluation metric under this basic prompting strategy, GPT-4o, despite its robustness, did not perform as well as a well-supported GPT-4.1.
\end{itemize}

These findings strongly support \textit{H1}, particularly for GPT-4.1, indicating that the explicit structural quality of legal text inputs is a critical determinant of its performance. GPT-4o appears largely resilient to such variations under basic prompting (but also lacks in overall performance).

\paragraph{Impact of Prompt Engineering and Structure Awareness (Hypothesis H2)}

Our second hypothesis (H2) proposed that explicitly informing the model about the nature of the structured input, particularly by shifting task instructions to the system prompt and adding a "structure awareness" cue, would further enhance its ability to leverage that structure. The results from Prompting Strategy S2 (System-Centric Task Definition with Structure Awareness) offer compelling support for this hypothesis, again most dramatically for GPT-4.1.

\textbf{Performance of GPT-4.1 under Strategy S2:}
\begin{itemize}
    \item Transitioning to Strategy S2 yielded significant additional performance gains for GPT-4.1 across all comparable structured input modalities.
    \begin{itemize}
        \item For $`CUAD_OrigTXT`$, accuracy increased from 67\% (S1) to 77\% (S2), an improvement of +10 percentage points.
        \item For $`AzureOCR_TXT`$, a less structured input, accuracy improved from 48\% (S1) to 56\% (S2), a gain of +8 percentage points. While still lagging behind better-structured inputs, this shows that improved prompting can partially mitigate the negative effects of poor input quality.
        \item Most notably, for $`GPT4oVision_MD`$ (Markdown), accuracy surged from 66\% (S1) to 79\% (S2), an uplift of +13 percentage points. This rendered Markdown the highest-performing input modality for GPT-4.1 under the optimized prompting conditions.
    \end{itemize}
    \item The superior performance of Markdown ($`GPT4oVision_MD`$ at 79\%) under Strategy S2, compared to even well-structured plain text ($`CUAD_OrigTXT`$ at 77\%), suggests that GPT-4.1, when explicitly primed by the system prompt to expect and process structured input, was able to more effectively leverage the explicit syntactic cues provided by Markdown.
\end{itemize}

\textbf{Performance of GPT-4o under Strategy S2:}
\begin{itemize}
    \item GPT-4o also showed improvements with Strategy S2, though more modest, consistent with its general robustness.
    \begin{itemize}
        \item For $`CUAD_OrigTXT`$, accuracy increased from 48\% (S1) to 50\% (S2) (+2pp).
        \item For $`AzureOCR_TXT`$, accuracy increased from 47\% (S1) to 51\% (S2) (+4pp).
        \item For $`GPT4oVision_MD`$, accuracy increased from 47\% (S1) to 52\% (S2) (+5pp), making it the best performing input for GPT-4o as well, albeit with a smaller margin of improvement.
    \end{itemize}
    \item While the gains were smaller, the pattern indicates that even the more resilient GPT-4o benefits to some degree from more explicit task definition and structure awareness cues in the system prompt.
\end{itemize}

These results strongly validate \textit{H2}. Strategic prompt engineering, specifically by relocating task instructions to the system prompt and including an advisory about the input's structured nature, significantly enhances model performance. For GPT-4.1, this enhancement is particularly pronounced and allows Markdown to demonstrate its full potential, outperforming other formats.

\section{Discussion}
\label{others}

The empirical findings reported in this study highlight the complex interactions among input text structure, prompt engineering strategies, and the performance of distinct Large Language Model architectures (GPT-4o and GPT-4.1) on a standardized legal question-answering task. These extend across the spectrum of legal practice, the design and implementation of legal technologies, and academic research on the reliability, effectiveness, and ethical oversight of AI in legal contexts.

\subsection{Implications for Legal Work}

The observed differential performance between GPT-4o and GPT-4.1, contingent upon variations in input text structure and the architecture of prompting strategies, offers several critical insights pertinent to the deployment and evolution of artificial intelligence within the legal profession.

\paragraph{Baseline Performance Dependency} The significant degradation in accuracy (a reduction of approximately 20 percentage points) when GPT-4.1 processed texts with compromised structure (such as those resulting from aggressive Regex-based cleaning or basic Azure OCR) compared to well-structured counterparts (like the original CUAD dataset texts or those parsed by GPT-4o Vision) under the initial, simpler prompting strategy, establishes a critical baseline. This finding underscores that for a considerable class of LLMs, meticulous data preprocessing is not merely an ancillary optimization but a foundational prerequisite for achieving even modest levels of reliable performance in complex legal tasks \cite{nawar2022, borchmann2024}.

Moreover, the adoption of structured formats provides fertile ground for the targeted reorganization of legal data, such as that contained in contracts. In addition to the accuracy gains demonstrated in this study, such formats enable clearer demarcation of legal units (e.g., sections or clauses), facilitating downstream tasks like retrieval, summarization, or clause classification. Looking ahead, lightweight markup languages like Markdown offer promising avenues for encoding logical separations and filters within legal documents, thereby unlocking additional functionality for diverse legal AI applications.

\paragraph{Meta-Cognitive Priming through Structure Awareness} The act of explicitly informing the model about the structured nature of its input (e.g., "You will receive structured Markdown") appears to function as a form of meta-cognitive priming. This allows the model to anticipate and potentially activate more specialized internal mechanisms for parsing and interpreting hierarchically organized information, leading to more effective utilization of the provided structural cues. This subtle yet powerful technique should be incorporated into advanced prompting playbooks. The differential response of GPT-4.1 and GPT-4o to the same prompting strategies underscores that there is no universal "optimal prompt." Prompting strategies must be tailored not only to the specific legal task but also to the particular LLM version being employed. This necessitates an iterative approach to prompt design, involving experimentation, rigorous evaluation, and continuous refinement.

\paragraph{Auditing Legacy Systems and Vendor Scrutiny} Many existing legal tech solutions may be built upon earlier LLM versions exhibiting sensitivities similar to GPT-4.1. Our findings suggest a need for organizations to potentially audit these systems, particularly examining their internal data ingestion and preprocessing pipelines. Furthermore, when procuring new LLM-powered tools, legal market participants should critically assess vendor claims regarding input handling and demand transparency concerning the model's sensitivity to data structure, ensuring that the tool can robustly handle the specific types and quality of documents prevalent in their workflows.

The substantial additional performance gains observed, particularly for GPT-4.1 (an uplift of ~10-13 percentage points), when transitioning from a user-centric to a system-centric prompting strategy that explicitly acknowledged input structure, elevates prompt engineering from a niche technical skill to a core competency in applied legal AI.

\paragraph{System Prompts as Anchors for Complex Legal Tasks} The efficacy of relocating core task instructions, output format specifications, and contextual priming to the system prompt is clearly demonstrated. This approach provides a more persistent and influential set of guidelines for the LLM, which is particularly beneficial for the multi-faceted and precise nature of legal question-answering and clause extraction tasks. Legal tech developers and sophisticated users should prioritize the strategic design of system prompts to fully harness the model's reasoning capabilities.

\paragraph{Strategic Implications for Document Management and Digitization} For law firms and legal departments, this translates into a strategic imperative to invest in and standardize high-quality document digitization and conversion processes. The reliance on rudimentary OCR technologies or indiscriminate text normalization routines—particularly those that inadvertently obliterate semantically relevant formatting such as paragraph breaks and indentations—can significantly impair the analytical capacity of downstream LLM applications. The superior textual output from advanced vision models, even when producing plain text, suggests a tangible return on investment for adopting state-of-the-art tools for converting scanned legal documents or poorly formatted electronic files into AI-ready structured data.

\paragraph{Beyond the "Newest is Best" Paradigm} The assumption that each successive LLM iteration will uniformly outperform its predecessors across all tasks and metrics is challenged by our findings. While GPT-4o's enhanced robustness to input variation is a valuable attribute, its lower absolute accuracy on this precise legal question-answering task (compared to an optimally prompted GPT-4.1) highlights the critical importance of task-specific benchmarking. Legal organizations must resist the allure of novelty for its own sake and instead base model selection on empirical evidence of performance for their specific use cases.
The observed performance differences may reflect distinct design emphases or fine-tuning objectives across model versions. These findings highlight the importance of aligning model selection with the specific requirements of the legal task in question.
The choice between a more sensitive, high-potential model and a more robust, potentially slightly less performant one should also be informed by an organization's available resources for data preprocessing and prompt engineering, as well as its risk tolerance. A firm with strong technical capabilities might extract superior value from a model like GPT-4.1, while another might prioritize the ease of deployment and broader applicability of a model like GPT-4o, accepting potential trade-offs in peak accuracy for specific narrow tasks.

\subsection{Limitations of the study}

This study provides empirical insights into how input structure and prompt design affect LLMs performance in legal tasks. These findings must be interpreted in light of key limitations stemming from our methodological choices and the evolving state of LLM research. These constraints limit the generalizability of our results and highlight important directions for future work.

This study focuses on two specific models—GPT-4.1 and GPT-4o—accessed via Microsoft Azure. While these represent prominent examples of advanced LLMs, they reflect only a small segment of the rapidly evolving landscape. Other models, such as Google’s Gemini, Anthropic’s Claude, and open-source architectures like Llama or Mistral, differ in training data, architecture, and fine-tuning approaches \footnote{For example, LLM Leaderboard (provider-agnostic) shows >15-point swings on MMLU, HumanEval and Latency across GPT-4o, Llama-3-70B, Mistral-Large and Gemini-Pro, underscoring heterogeneous trade-offs. \cite{llmleaderboard}}. As such, the input sensitivities observed—particularly GPT-4.1’s heightened responsiveness compared to GPT-4o’s robustness—and the prompt strategies evaluated may not generalize across models \footnote{For example, OpenAI’s GPT-4o system card notes robustness gaps outside its TTS test-set and calls for “task-specific validation before deployment.” \cite{openaigpt4o}}. These findings underscore the need for caution in drawing broad conclusions and point to the importance of future cross-architectural studies to assess the generality of these effects.

This study focuses on a single, yet complex and critical legal task: clause identification and verbatim extraction in a question-answering setting, using an excerpt of the Contract Understanding Atticus Dataset (CUAD). The domain of legal informatics spans a much wider range of tasks, including statutory and case law reasoning, judicial opinion summarization, legal drafting, outcome prediction, and evidentiary analysis. The effectiveness of input structuring and prompting strategies is likely to vary across these tasks, depending on their cognitive complexity. For example, tasks involving abstractive reasoning or generation may require different approaches than the extractive task explored here. Therefore, the direct applicability of our conclusions to the entirety of legal AI applications must be approached with circumspection.

A central component of our methodology involved the generation of structured inputs, particularly Markdown, from hypothetical image-based PDFs using the vision capabilities of GPT-4o. While using vision models for document understanding is a promising and increasingly viable approach, the specifics of the vision-to-Markdown pipeline—such as prompts, operational steps, and internal mechanisms—were not explicitly analyzed as variables in this study \footnote{We used the github library vision-parse with its standard configuration. \cite{visionparse}}. The resulting Markdown’s quality and accuracy depend on the vision model’s capabilities and the instructions provided. Uncontrolled variations in this upstream conversion process could introduce confounders, influencing the downstream LLM's performance in ways not fully attributable to the Markdown format per se. Future work should systematically explore methods for reliably transforming unstructured or image-based legal documents into high-fidelity structured representations (see more in Chapter 4.3).

We evaluated performance using exact-match accuracy\footnote{For more information reference to Deepset Guide. \cite{deepsetmetrics}} for a bipartite output: a correct “Yes/No” decision and a verbatim clause citation. While this metric aligns with the precision often required in legal contexts, it imposes a strict evaluation standard that may overlook partial understanding, semantically equivalent responses, or correct identification of legal concepts without exact text matches. Alternative metrics—such as semantic similarity scores (e.g., ROUGE, BERTScore)\cite{zhang2024}, \textit{F1} for concept recognition \cite{f1score}, or human judgment—could provide a more nuanced assessment, especially for tasks where minor deviations still hold legal relevance. Relying solely on exact-match may have understated certain model capabilities \cite{chen2019}. 

This study primarily focused on the impact of relatively explicit and macroscopic structural elements, such as paragraph breaks, section demarcations (implicitly conveyed through formatting or explicitly encoded in Markdown headings), and list structures. Legal documents, however, are often characterized by far more intricate and subtle structural complexities (see e.g. \cite{euguidelines2025}). These include deeply nested clausal hierarchies, complex inter-document and intra-document cross-referencing schemes, the precise definitional scope and application of terms, and the logical flow of obligations and conditions. The optimal methods for representing these more granular and relational structural features to LLMs, and the extent to which current models can effectively leverage such sophisticated information, remain largely unexplored territories. Our findings concerning basic structure may not fully extend to these more challenging aspects of legal document architecture.

Models are subject to unannounced iterative updates, changes in training data cutoffs, and modifications to their fine-tuning or alignment procedures. This inherent fluidity can pose challenges for exact replication and direct comparison of results across different time points or even different API endpoints. While we endeavored to define our model targets as clearly as possible within these constraints, the potential for subtle, unacknowledged model drift cannot be entirely discounted and is a systemic challenge in research involving proprietary, rapidly evolving LLMs.

Although our study systematically compared two distinct prompting strategies (user-centric vs. system-centric with structure awareness), the vast and multi-dimensional parameter space of prompt engineering was, by necessity, not exhaustively traversed. Advanced prompting techniques—such as the inclusion of few-shot exemplars \cite{brown2020}, the elicitation of chain-of-thought \cite{wei2023} or step-by-step reasoning \cite{lightman2023}, the use of self-consistency methods \cite{wang2023}, or more elaborate role-playing scenarios—were not systematically applied to each input modality and model combination. It remains plausible that more highly specialized or refined prompting strategies, tailored with even greater precision, could further modulate the observed performance characteristics, potentially altering the relative efficacy of the input formats or eliciting different sensitivities from the models. Our study provides a valuable signal regarding broad prompt architecture, but does not claim to have identified the absolute optimal prompt for any given condition.

\subsection{Further research}

The findings and limitations of this study point to several important directions for future research, aimed at broadening the scope, increasing analytical depth, and addressing emerging challenges in applying Large Language Models to legal tasks. These avenues are essential for advancing both understanding and practical deployment in this evolving domain.

A primary avenue for future work involves extending the comparative analysis to a wider array of LLM architectures and providers. Replicating the core experimental design—examining sensitivity to input structure and prompt engineering—across diverse models (including leading open-source alternatives and models from different commercial vendors) would be invaluable. Such research could identify whether the observed patterns of sensitivity (e.g., GPT-4.1-like dependence on structure vs. GPT-4o-like robustness) are idiosyncratic to specific model lineages or represent more generalizable characteristics of certain architectural classes or training paradigms. This would contribute to a more comprehensive "Consumer Reports"-style understanding of LLM behaviors for legal applications.

Future studies should systematically investigate the impact of input structure and prompting on a wide range of legal tasks, not just contract review. This includes, but is not limited to:
\begin{itemize}
    \item Legal Reasoning and Argumentation: How does the structured presentation of case law, statutes, and factual scenarios influence an LLM's ability to construct coherent legal arguments or predict judicial outcomes?
    \item Document Summarization and Synthesis: What input formats best enable LLMs to generate accurate and concise summaries of lengthy legal texts, such as depositions or judicial opinions, while preserving critical nuances?
    \item Legal Document Drafting: How can structured templates or Markdown-based outlines be used to guide LLMs in drafting more reliable and compliant legal documents (e.g., motions, contracts, wills)?
    \item Employing a wider range of legal datasets, each with its unique structural properties and task demands, will be crucial for developing a more holistic understanding.
\end{itemize}

Research is needed to understand how LLMs can effectively process and utilize more complex and subtle structural features inherent in legal texts. This includes:
\begin{itemize}
    \item Investigating optimal ways to represent and make salient intra-document and inter-document cross-references, nested clausal dependencies, definitional hierarchies, and the scope of legal provisions.
    \item Developing techniques that enable LLMs to reason over these complex relational structures, rather than just processing linear text.
    \item Exploring the utility of knowledge graphs or other structured knowledge representations, potentially generated from or used in conjunction with LLMs, to model these intricate legal relationships.
\end{itemize}

Future research should move beyond exact-match accuracy by incorporating a broader set of evaluation metrics. Semantic similarity measures (e.g.ROUGE\cite{zhang2024} or BLEU\cite{bleuscore}) can capture fidelity in outputs like summaries or paraphrased clauses, even when they diverge from exact source wording. Equally important is the development of factuality and faithfulness metrics tailored to legal contexts, enabling rigorous assessment of whether outputs accurately reflect source documents and helping to detect hallucinations\cite{ragas2025}. Metrics for evaluating logical coherence and legal reasoning quality are also critical. Additionally, task-specific human evaluation frameworks, grounded in detailed rubrics and expert legal judgment, will be essential for assessing real-world utility and correctness. Approaches such as LLM-as-a-judge\cite{llmasajudge}, increasingly common in current evaluation practice, offer promising directions, as does the development of structured legal benchmarking protocols—potentially rule-based—for systematic performance comparison across models and tasks.

Future work must also consider the ethical dimensions of structuring legal data for LLMs. For instance, does the process of converting diverse, unstructured legal narratives into standardized structured formats inadvertently filter out important contextual nuances or amplify existing biases present in the source material? Research is needed to develop best practices for equitable and responsible data structuring in the legal domain.

\section{Conclusion}

This study examined the often-overlooked impact of input structure and prompt design on the performance of GPT-4o and GPT-4.1 in legal question-answering. The results highlight a complex interaction between model architecture, data representation, and prompting strategies, challenging assumptions about LLMs generality and emphasizing the need for careful, context-sensitive deployment in legal applications.

The core results demonstrate a notable divergence in model sensitivity. While GPT-4o exhibited considerable robustness to variations in input text structure, its absolute accuracy on the defined task, under the tested conditions, was surpassed by GPT-4.1. Conversely, GPT-4.1 proved highly susceptible to the quality of input formatting; its performance on poorly structured texts was markedly inferior, yet improved substantially (by approximately 20 percentage points) when provided with well-structured inputs, whether derived from original well-formatted CUAD texts or generated by advanced vision models like GPT-4o Vision. Crucially, a further significant performance uplift for GPT-4.1 (an additional ~10-13 percentage points) was achieved by relocating task instructions to the system prompt and explicitly apprising the model of the structured nature of its input. Under these optimized conditions, Markdown emerged as the most effective input format for GPT-4.1, achieving an overall exact-match accuracy of 79\%.

The findings support the thesis that for many LLMs—including advanced models like GPT-4.1—input structuring and prompt design are critical to performance, not peripheral details. While newer models such as GPT-4o show greater robustness, they still benefit from tailored optimization and do not consistently outperform predecessors across all tasks without it. Small variations in input formatting or phrasing can lead to significant differences in outcomes, a factor of particular importance in high-stakes legal contexts.

These results have practical implications for legal professionals, developers, and researchers. They highlight the need for robust data preparation pipelines, advanced prompt engineering skills, and rigorous task-specific benchmarking. This research reinforces the view of LLMs as assistive tools whose effectiveness depends heavily on expert-guided use, rather than as fully autonomous systems.

Even under optimized conditions, the peak accuracy observed in this study (79\%) falls short of the reliability required in many legal contexts. This underscores the role of LLMs as assistive tools rather than autonomous legal decision-makers. The “human-in-the-loop” model remains not a temporary safeguard, but a long-term necessity. Legal professionals must maintain realistic expectations and ensure transparent communication about AI’s role and limitations. Crucially, beyond overall accuracy, organizations must systematically analyze LLM errors—their types, frequency, and potential impact—to refine models, strengthen human oversight, and manage the risks of AI-assisted legal practice.

As large language models become more prevalent in legal practice, there is an urgent need for critical evaluation, methodological robustness and an in-depth understanding of the factors that influence their behaviour. This study contributes to that effort, showing that progress in legal AI depends not only on model advancements but also on how carefully we design, guide, and evaluate their use. Such nuanced understanding is essential for the responsible and effective integration of AI into legal work.

\section*{Acknowledgements}

We would like to thank Daniel Mentjukov for providing the pre-processed dataset, initial evaluation framework, on which this study builds. His contributions were instrumental to the development of our methodology. We are also grateful to Gereon Abendroth and Marc Ohrendorf for their valuable input and thought-provoking discussions, which helped shape the direction of this work. Our thanks also go to the entire Osborne Clarke Germany team for their mutual support across all projects, and for challenging, explaining, and helping us refine studies like this one. This research was conducted within the context of Osborne Clarke Solutions Germany, the legal tech unit of Osborne Clarke Germany.

\bibliography{references}

\begin{thebibliography}{10}

\bibitem{llmleaderboard}
{Artificial Analysis}.
\newblock Llm leaderboard — compare gpt-4o, llama 3, mistral, gemini \& other
  models.
\newblock \url{https://artificialanalysis.ai/leaderboards/models}, 2025.
\newblock Accessed 2025-05-15.

\bibitem{bender2021}
Emily~M. Bender, Timnit Gebru, Angelina McMillan-Major, and S.~Shmitchell.
\newblock On the dangers of stochastic parrots: Can language models be too big?
\newblock In {\em Proceedings of the 2021 ACM Conference on Fairness,
  Accountability, and Transparency}, pages 610--623, Virtual Event, Canada,
  March 2021. ACM.

\bibitem{borchmann2024}
Łukasz Borchmann.
\newblock Notes on applicability of gpt-4 to document understanding, May 2024.

\bibitem{visionparse}
Arun Brahma.
\newblock Vision-parse: Parse pdfs into markdown using vision llms.
\newblock \url{https://github.com/iamarunbrahma/vision-parse}, December 2024.
\newblock Python software, accessed 2025-05-14.

\bibitem{brown2020}
Tom~B. Brown et~al.
\newblock Language models are few-shot learners, July 2020.

\bibitem{chen2019}
Allen Chen, Gabriel Stanovsky, Sameer Singh, and Matt Gardner.
\newblock Evaluating question answering evaluation.
\newblock In {\em Proceedings of the 2nd Workshop on Machine Reading for
  Question Answering}, pages 119--124, Hong Kong, China, 2019. Association for
  Computational Linguistics.

\bibitem{chen2025}
Zeyu Chen et~al.
\newblock Mdeval: Evaluating and enhancing markdown awareness in large language
  models, January 2025.

\bibitem{f1score}
{Data Science Dojo}.
\newblock What is f1 score? an essential metric in llm evaluation.
\newblock \url{https://datasciencedojo.com/blog/understanding-f1-score/}, 2025.
\newblock Accessed 2025-05-15.

\bibitem{deepsetmetrics}
{deepset}.
\newblock Metrics to evaluate a question answering system.
\newblock
  \url{https://www.deepset.ai/blog/metrics-to-evaluate-a-question-answering-system},
  2025.
\newblock Accessed 2025-05-15.

\bibitem{gelbach2021}
David~Freeman Engstrom and Jonah~B. Gelbach.
\newblock Legal tech, civil procedure, and the future of adversarialism, 2021.
\newblock Unpublished manuscript.

\bibitem{ragas2025}
Sreyan Es, Jack James, Luis Espinosa-Anke, and Steven Schockaert.
\newblock Ragas: Automated evaluation of retrieval augmented generation, April
  2025.

\bibitem{legalbench2023}
Nilay Guha et~al.
\newblock Legalbench: A collaboratively built benchmark for measuring legal
  reasoning in large language models, August 2023.

\bibitem{hamad2016}
Khaled Hamad and Mehmet Kaya.
\newblock A detailed analysis of optical character recognition technology.
\newblock {\em International Journal of Applied Mathematics, Electronics and
  Computers}, 4(Special Issue-1):244--244, December 2016.

\bibitem{han2025}
Yujia Han, Yan Wu, and Jacob Willard.
\newblock Effect of selection format on llm performance, March 2025.

\bibitem{he2024}
Jinyuan He, Mansi Rungta, Dylan Koleczek, Amandeep Sekhon, Fangxiang~X. Wang,
  and Samira Hasan.
\newblock Does prompt formatting have any impact on llm performance?, November
  2024.

\bibitem{hegel2021}
Alexander Hegel, Mohammed Shah, Graham Peaslee, Benjamin Roof, and Ehab Elwany.
\newblock The law of large documents: Understanding the structure of legal
  contracts using visual cues, July 2021.

\bibitem{hegghammer2022}
Thomas Hegghammer.
\newblock Ocr with tesseract, amazon textract, and google document ai: a
  benchmarking experiment.
\newblock {\em Journal of Computational Social Science}, 5(1):861--882, May
  2022.

\bibitem{hendrycks2021}
Dan Hendrycks, Collin Burns, Anya Chen, and Steven Ball.
\newblock Cuad: An expert-annotated nlp dataset for legal contract review,
  November 2021.

\bibitem{lai2023}
Jia Lai, Wei Gan, Jiahao Wu, Ziqi Qi, and Philip~S. Yu.
\newblock Large language models in law: A survey, November 2023.

\bibitem{liang2023}
Percy Liang et~al.
\newblock Holistic evaluation of language models, October 2023.

\bibitem{lightman2023}
Howard Lightman et~al.
\newblock Let’s verify step by step, May 2023.

\bibitem{martin2024}
Lucy Martin, Neil Whitehouse, Sandy Yiu, Lewis Catterson, and Rishad Perera.
\newblock Better call gpt, comparing large language models against lawyers,
  January 2024.

\bibitem{microsoftOCR}
Microsoft.
\newblock Azure ai vision with ocr and ai.
\newblock
  \url{https://azure.microsoft.com/en-us/products/ai-services/ai-vision}, 2025.
\newblock Accessed 2025-05-14.

\bibitem{nawar2022}
Anas Nawar, Md~Rakib, Sadik~A. Hai, and Sufiyan Haq.
\newblock An open source contractual language understanding application using
  machine learning.
\newblock In {\em Proceedings of the First Workshop on Language Technology and
  Resources for a Fair, Inclusive, and Safe Society within the 13th Language
  Resources and Evaluation Conference}, pages 42--50, Marseille, France, June
  2022. European Language Resources Association.

\bibitem{nguyen2022}
Truc T.~H. Nguyen, Adam Jatowt, Matthieu Coustaty, and Antoine Doucet.
\newblock Survey of post-ocr processing approaches.
\newblock {\em ACM Computing Surveys}, 54(6):1--37, July 2022.

\bibitem{openai2025}
OpenAI.
\newblock Gpt-4.1 prompting guide\,|\,openai cookbook.
\newblock \url{https://cookbook.openai.com/examples/gpt4-1_prompting_guide},
  2025.
\newblock Accessed 2025-05-13.

\bibitem{openaigpt4o}
OpenAI.
\newblock Gpt-4o system card.
\newblock \url{https://openai.com/index/gpt-4o-system-card/}, 2025.
\newblock Accessed 2025-05-14.

\bibitem{openaigpt4.1}
OpenAI.
\newblock Introducing gpt-4.1 in the api.
\newblock \url{https://openai.com/index/gpt-4-1/}, 2025.
\newblock Accessed 2025-05-14.

\bibitem{opengpt42024}
OpenAI et~al.
\newblock Gpt-4 technical report, March 2024.

\bibitem{bleuscore}
Kishore Papineni, Salim Roukos, Todd Ward, and Wei-Jing Zhu.
\newblock Bleu: A method for automatic evaluation of machine translation.
\newblock In {\em Proceedings of the 40th Annual Meeting of the Association for
  Computational Linguistics}, page 311, Philadelphia, Pennsylvania, 2002.
  Association for Computational Linguistics.

\bibitem{pipitone2024}
Niccolò Pipitone and Ghadeer~H. Alami.
\newblock Legalbench-rag: A benchmark for retrieval-augmented generation in the
  legal domain, August 2024.

\bibitem{polo2024}
Francisco~M. Polo et~al.
\newblock Efficient multi-prompt evaluation of llms, October 2024.

\bibitem{euguidelines2025}
{Publications Office of the EU}.
\newblock 2.\ structure of a legal act — interinstitutional style guide.
\newblock
  \url{https://style-guide.europa.eu/en/content/-/isg/topic?identifier=2-structure-legal-act},
  2025.
\newblock Accessed 2025-05-14.

\bibitem{romoser2023}
Jordan Romoser.
\newblock No, ruth bader ginsburg did not dissent in obergefell — and other
  things chatgpt gets wrong about the supreme court.
\newblock
  \url{https://www.scotusblog.com/2023/01/no-ruth-bader-ginsburg-did-not-dissent-in-obergefell-and-other-things-chatgpt-gets-wrong-about-the-supreme-court/},
  2023.
\newblock SCOTUSblog, accessed 2025-05-13.

\bibitem{sahoo2025}
Priyabrata Sahoo, Ayush~K. Singh, Souvik Saha, Vaibhav Jain, Subhankar Mondal,
  and Aditya Chadha.
\newblock A systematic survey of prompt engineering in large language models:
  Techniques and applications, March 2025.

\bibitem{schulhoff2025}
Samuel Schulhoff et~al.
\newblock The prompt report: A systematic survey of prompt engineering
  techniques, February 2025.

\bibitem{shen2025}
Jiawei Shen et~al.
\newblock A law reasoning benchmark for llm with tree-organized structures
  including factum probandum, evidence and experiences, March 2025.

\bibitem{sonoda2024}
Yuji Sonoda et~al.
\newblock Structured clinical reasoning prompt enhances llm’s diagnostic
  capabilities in diagnosis please quiz cases.
\newblock {\em Radiology and Imaging}, September 2024.
\newblock Preprint.

\bibitem{surden2020}
Harry Surden.
\newblock Ethics of ai in law: Basic questions.
\newblock In Markus~Dirk Dubber, Frank Pasquale, and Sunit Das, editors, {\em
  The Oxford Handbook of Ethics of AI}, pages 719--736. Oxford University
  Press, 2020.

\bibitem{tan2024}
Xiaoming Tan et~al.
\newblock Struct-x: Enhancing large language models reasoning with structured
  data, July 2024.

\bibitem{volokh2023}
Eugene Volokh.
\newblock Chatgpt coming to court, by way of self-represented litigants.
\newblock
  \url{https://reason.com/volokh/2023/05/27/chatgpt-coming-to-court-by-way-of-self-represented-litigants/},
  2023.
\newblock Reason.com, accessed 2025-05-13.

\bibitem{wang2024}
Junlin Wang et~al.
\newblock Legal evaluations and challenges of large language models, November
  2024.

\bibitem{wang2023}
Shuo Wang et~al.
\newblock Maud: An expert-annotated legal nlp dataset for merger agreement
  understanding, 2023.

\bibitem{wei2023}
Jason Wei et~al.
\newblock Chain-of-thought prompting elicits reasoning in large language
  models, January 2023.

\bibitem{weiser2025}
Benjamin Weiser.
\newblock Here’s what happens when your lawyer uses chatgpt.
\newblock
  \url{https://www.nytimes.com/2023/05/27/nyregion/avianca-airline-lawsuit-chatgpt.html},
  May 2023.
\newblock The New York Times, accessed 2025-05-13.

\bibitem{zhang2024}
Ming Zhang, Chengpeng Li, Min Wan, Xiaoxuan Zhang, and Qian Zhao.
\newblock Rouge-sem: Better evaluation of summarization using rouge combined
  with semantics.
\newblock {\em Expert Systems with Applications}, 237:121364, March 2024.

\bibitem{llmasajudge}
Liangliang Zheng et~al.
\newblock Judging llm-as-a-judge with mt-bench and chatbot arena, December
  2023.

\bibitem{zhu2024}
Kai Zhu et~al.
\newblock Promptbench: A unified library for evaluation of large language
  models, August 2024.

\end{thebibliography}

\newpage

\appendix
\section{\\ Used questions and question template as part of CUAD} \label{appendix:A}

\begin{lstlisting}
question_templates = [
    "Does the {contract} contract contain an exception or carve-out to any competitive restriction, allowing certain competitive activities or partnerships that would otherwise be prohibited, and if so, does it specify the activities or partnerships allowed?",
    "Does the {contract} contract include a clause that restricts a party from competing or engaging in a business activity that competes with the other party, and if so, does it specify a time frame or geographic area for this restriction?",
    "Does the {contract} contract grant one party exclusive rights to provide certain goods or services, and if so, does it also prevent that party from entering into similar agreements with third parties?",
    "Is there a provision in the {contract} contract that prohibits a party from soliciting or doing business with the customers or clients of the other party, and if so, does it specify a duration for this prohibition?"
]
\end{lstlisting}

\section{\\Example Result Question Answering} \label{appendix:B}

\begin{figure}[htbp]
  \centering
  \includegraphics[width=\linewidth]{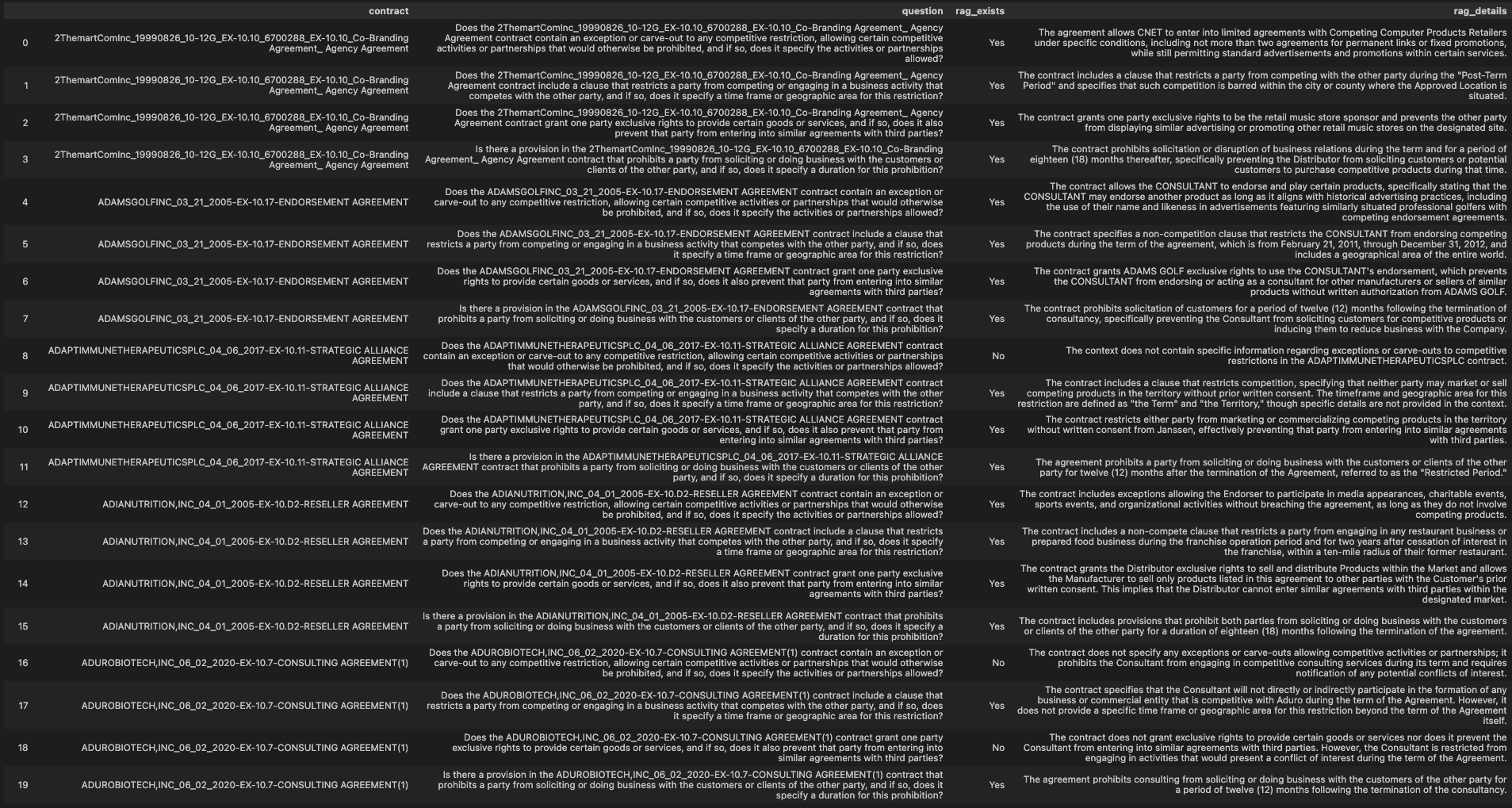}
  \caption{Results after answering the questions and extracting the relevant clauses for answering (excerpt).}
  \label{fig:answer_results}
\end{figure}

\section{\\ Vision-Parse Custom Prompt} \label{appendix:C}

\begin{lstlisting}
ZERO_SHOT ="""
Your task is to read the given contract and convert its content into clean, well-structured Markdown format.
You are a legal document structure analyst.

Important: The only reason for structuring markdown is to enhance legal coherence and legal logic.

<instructions>
- Detect and preserve all headings using #, ##, ### based on their visual hierarchy (font size, boldness).
- Structure paragraphs clearly, utilizing bullet points, numbering, and formatting (bold, italic) when applicable.
- Convert all tables into proper Markdown table format with headers and aligned rows.
- Ignore dividers.
- Ignore irrelevant elements like footers, page numbers, and watermarks.
- Maintain logical reading order.
</instructions>

<output>
Output only the structured Markdown content. Do not explain or add comments.
</output>
"""
\end{lstlisting}

\section{\\Question Distribution of used dataset from CUAD} \label{appendix:D}

\begin{figure}[htbp]
  \centering
  \includegraphics[width=\linewidth]{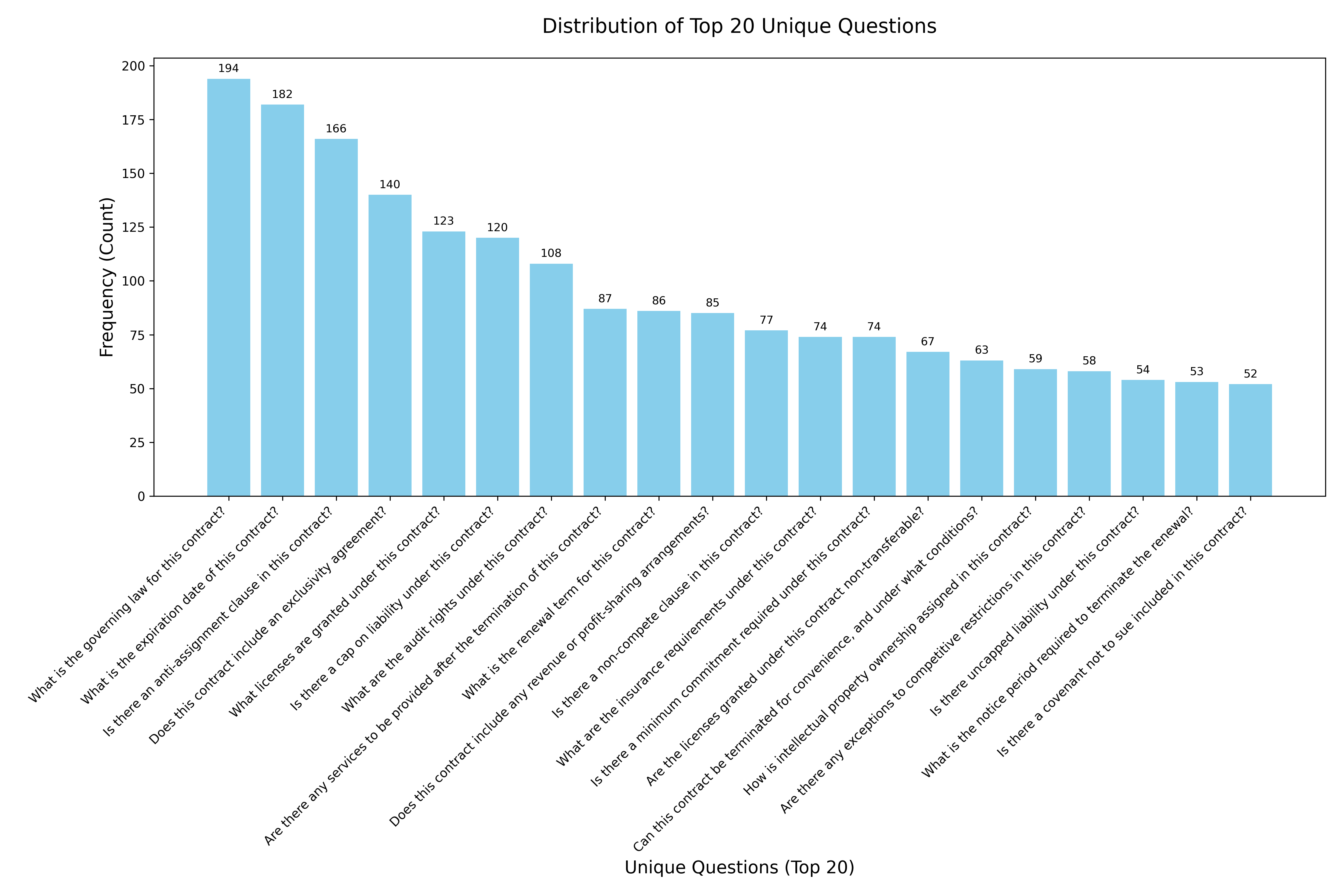}
  \caption{Top 20 questions distribution from the used excerpt of the CUAD dataset.}
  \label{fig:question_distribution}
\end{figure}

\begin{table}[htbp]
  \centering
  \caption{Unique question count of all questions derived from the excerpt of the CUAD dataset.}
  \begin{tabularx}{\textwidth}{@{}Xr@{}}
    \toprule
    \textbf{Question} & \textbf{Count} \\
    \midrule
    Are the licenses granted under this contract non-transferable? & 67 \\
    Are there any exceptions to competitive restrictions in this contract? & 58 \\
    Does this contract include an exclusivity agreement? & 140 \\
    Is there a cap on liability under this contract? & 120 \\
    Is there an anti-assignment clause in this contract? & 166 \\
    What are the insurance requirements under this contract? & 74 \\
    What is the expiration date of this contract? & 182 \\
    What is the governing law for this contract? & 194 \\
    What licenses are granted under this contract? & 123 \\
    Can this contract be terminated for convenience, and under what conditions? & 63 \\
    Does this contract include any right of first refusal, right of first offer, or right of first negotiation? & 48 \\
    What are the audit rights under this contract? & 108 \\
    Are there any services to be provided after the termination of this contract? & 87 \\
    Does the licensee's affiliates have any licensing rights under this contract? & 33 \\
    Does this contract include any revenue or profit-sharing arrangements? & 85 \\
    Is there a covenant not to sue included in this contract? & 52 \\
    Is there a minimum commitment required under this contract? & 74 \\
    Is there a non-compete clause in this contract? & 77 \\
    What is the notice period required to terminate the renewal? & 53 \\
    What is the renewal term for this contract? & 86 \\
    Is there a clause preventing the solicitation of customers in this contract? & 17 \\
    What happens in the event of a change of control of one of the parties in this contract? & 49 \\
    What is the duration of any warranties provided in this contract? & 35 \\
    Are there any third-party beneficiaries designated in this contract? & 12 \\
    Are any of the licenses granted under this contract irrevocable or perpetual? & 28 \\
    How is intellectual property ownership assigned in this contract? & 59 \\
    Is there a clause preventing the solicitation of employees in this contract? & 25 \\
    Does this contract include an unlimited or all-you-can-eat license? & 8 \\
    Does this contract include any volume restrictions? & 49 \\
    Is there uncapped liability under this contract? & 54 \\
    Does this contract provide for joint intellectual property ownership? & 25 \\
    Does the licensor's affiliates have any licensing rights under this contract? & 8 \\
    Is there a most favored nation clause in this contract? & 18 \\
    Are there any price restrictions or controls specified in this contract? & 7 \\
    Is there a non-disparagement clause in this contract? & 12 \\
    \bottomrule
  \end{tabularx}
  \label{tab:cuad_question_counts}
\end{table}

\newpage
\section{\\Derived question for ground truth from CUAD dataset (excerpt)} \label{appendix:E}

\begin{figure}[htbp]
  \centering
  \includegraphics[width=\linewidth]{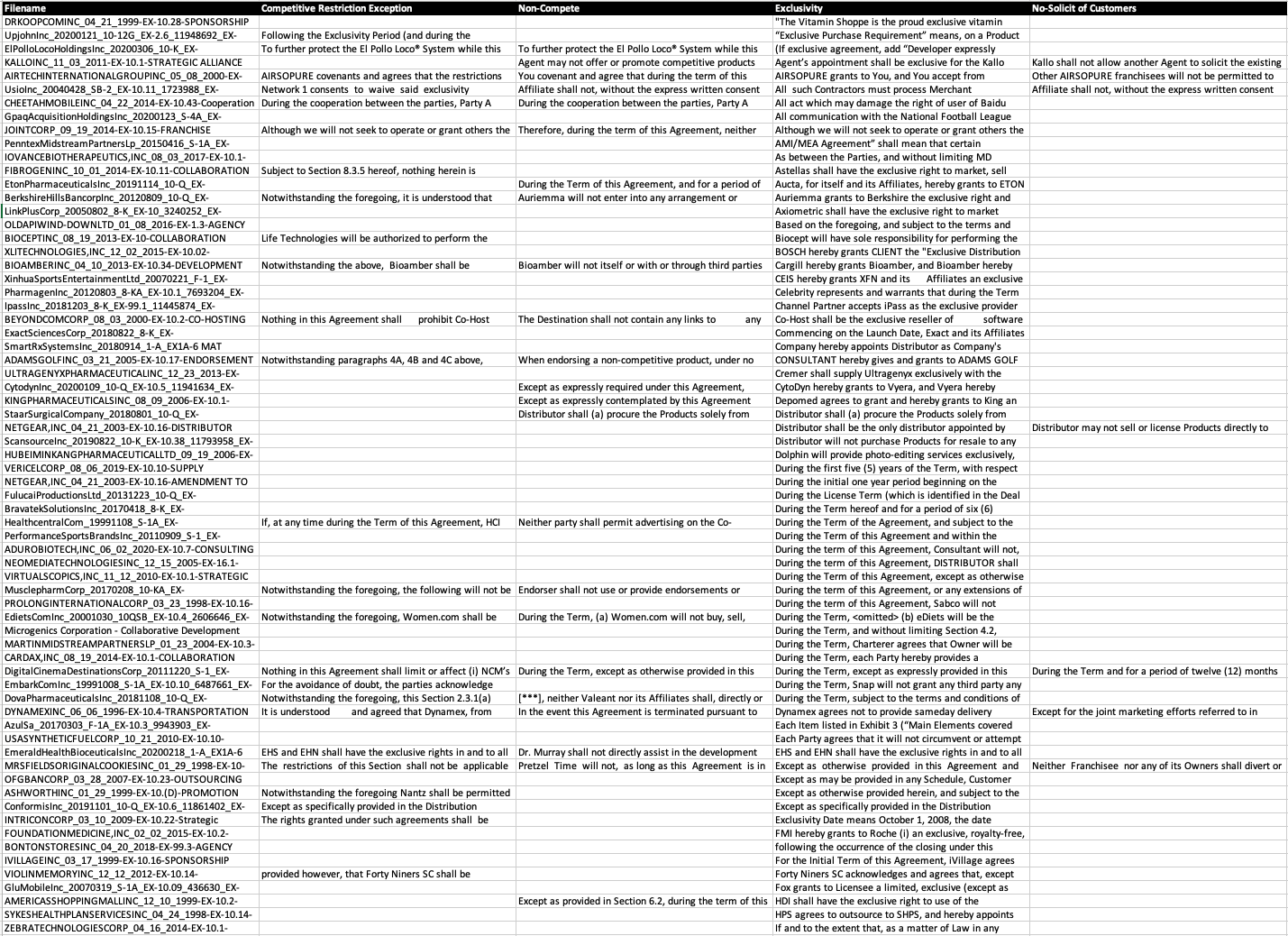}
  \caption{Building ground truth from CUAD dataset}
  \label{fig:question_distribution}
\end{figure}

\end{document}